\documentclass[conference]{IEEEtran}
\IEEEoverridecommandlockouts
\usepackage{amsmath,amssymb,amsfonts}
\usepackage{algorithmic}
\usepackage{graphicx}
\usepackage{textcomp}
\usepackage{xcolor}
\usepackage{todonotes}
\usepackage[numbers]{natbib}
\usepackage{graphicx}
\usepackage{hyperref}
\usepackage{subfigure}
\usepackage{comment}
\usepackage{tikz}

\usepackage{tikz}
\usepackage{tikz-3dplot}
\usepackage{tkz-euclide}
\usetikzlibrary{calc,patterns,angles,quotes}

\usepackage{tikz-3dplot}
\definecolor{dark-red}{rgb}{0.4,0.15,0.15}
\definecolor{dark-blue}{rgb}{0.15,0.15,0.6}
\definecolor{medium-blue}{rgb}{0,0,0.5}
\long\def\invis#1{}
\long\def\greg#1{{{\bf\color{red} Greg: #1}}}

\newcommand\gderror[1]{
   \typeout{--------------------------------------------------------------------}
   \typeout{------- #1 ---------}
   \typeout{--------------------------------------------------------------------}
   {\bf #1}
}
\newcounter{gdTmp} 
\newcounter{gdLastCount}
\newcommand\maxpage[2][Error]{  
\ifnum\value{page}>#2
    \gderror{On page {\thepage} we are past page #2 (too long).   #1 }
\else\fi
\setcounter{gdLastCount}{\value{page}} 
}
\newcommand\maxpageSinceLast[2][Error]{  
\ifnum \numexpr \value{page} - \value{gdLastCount}\relax>#2
    \gderror{Exceeds max length #2 pages. Page \thepage: #1}
\thepage\else\fi
\setcounter{gdLastCount}{\value{page}} 
}
\long\def\greg#1{{{\bf\color{red} Greg: #1}}}

\def\BibTeX{{\rm B\kern-.05em{\sc i\kern-.025em b}\kern-.08em
    T\kern-.1667em\lower.7ex\hbox{E}\kern-.125emX}}
\begin{document}

\makeatletter
\newcommand{\newlineauthors}{%
  \end{@IEEEauthorhalign}\hfill\mbox{}\par
  \mbox{}\hfill\begin{@IEEEauthorhalign}
}
\makeatother

\title{\fontsize{20pt}{20pt}\selectfont Scalable Aerial GNSS Localization for Marine Robots\\[0.5em]
{\fontsize{12pt}{14pt}\selectfont Shuo Wen$^{1*}$, Edwin Meriaux$^{1*}$, Mariana Sosa Guzmán$^1$, Charlotte Morissette$^1$,\\ Chloe Si$^2$, Bobak Baghi$^3$, Gregory Dudek$^4$}

\thanks{ Paper accepted at International Conference on Robotics and Automation 2025 Workshop Robots in the Wild \\
$^1$Center for Intelligent Machines (CIM) Laboratory, McGill University, Montreal, Canada \{Shuo.Wen, Edwin.Meriaux, Mariana.Sosaguzman, Charlotte.Morissette\}@mail.mcgill.ca \\
$^2$ Department of Mathematics and Statistics, McGill University, Montreal, Canada \{Chuqiao.Si\}@mail.mcgill.ca\\
$^3$ Independent Researcher \{Bobak.Hamed-Baghi\}@mail.mcgill.ca\\
$^4$ Center for Intelligent Machines (CIM) Laboratory, McGill University, Montreal, Canada \{Gregory.Dudek\}@mcgill.ca \\

$^*$ Co-first authors with equal contributions.

}
}

\maketitle
\begin{abstract}
Accurate localization is crucial for water robotics, yet traditional onboard Global Navigation Satellite System (GNSS) approaches are difficult or ineffective due to signal reflection on the water’s surface and its high cost of aquatic GNSS receivers. Existing approaches, such as inertial navigation, Doppler Velocity Loggers (DVL), SLAM, and acoustic-based methods, face challenges like error accumulation and high computational complexity. Therefore, a more efficient and scalable solution remains necessary. This paper proposes an alternative approach that leverages an aerial drone equipped with GNSS localization to track and localize a marine robot once it is near the surface of the water. Our results show that this novel adaptation enables accurate single and multi-robot marine robot localization.  
\end{abstract}

\begin{IEEEkeywords}
Localization, GNSS, Marine Robotics, Field Robotics, Drones 
\end{IEEEkeywords}

\section{Introduction}

In this paper, we explore the use of an aerial drone equipped with GNSS localization to track and localize a marine robot when it is near the surface, presenting a novel adaptation of this technique for multi-robot localization in marine environments.

Robot localization typically involves the use of a Global Navigation Satellite System (GNSS) such as the Global Positioning System (GPS) that estimates the position of a robot through the use of various satellites. Extending this solution to marine robots is challenging due to the limitations of GNSS signal propagation. The signals cannot be received underwater because seawater is a conductive medium, and the signals are often of low signal quality very near the surface in choppy water. As a consequence, marine robots equipped with GNSS receivers must surface, often for extended intervals,  to connect to the positioning system. Of course, a GNSS receiver needs an external antenna to receive the required signals, which has disadvantages for a sealed vehicle. These factors have led many researchers to explore marine robot tracking without the use of GNSS receivers. 

An extensive body of work has explored the use of inertial sensors in combination with Doppler Velocity Loggers (DVL) for marine localizations \cite{turbulent_ocean}, \cite{artic_dvl}. Others have explored the use of acoustic-based techniques \cite{lbl3}, \cite{lbl2}, \cite{lbl1},  or Simultaneous Localization and Mapping (SLAM) \cite{slam_fail1} ones. Although these approaches present possible solutions to the marine localization problem, each suffers from a different limitation, such as error accumulation or high computational complexity. Hence, an efficient and scalable solution is still needed.

\begin{figure}[t]
    \centering
    \includegraphics[width=0.45\textwidth]{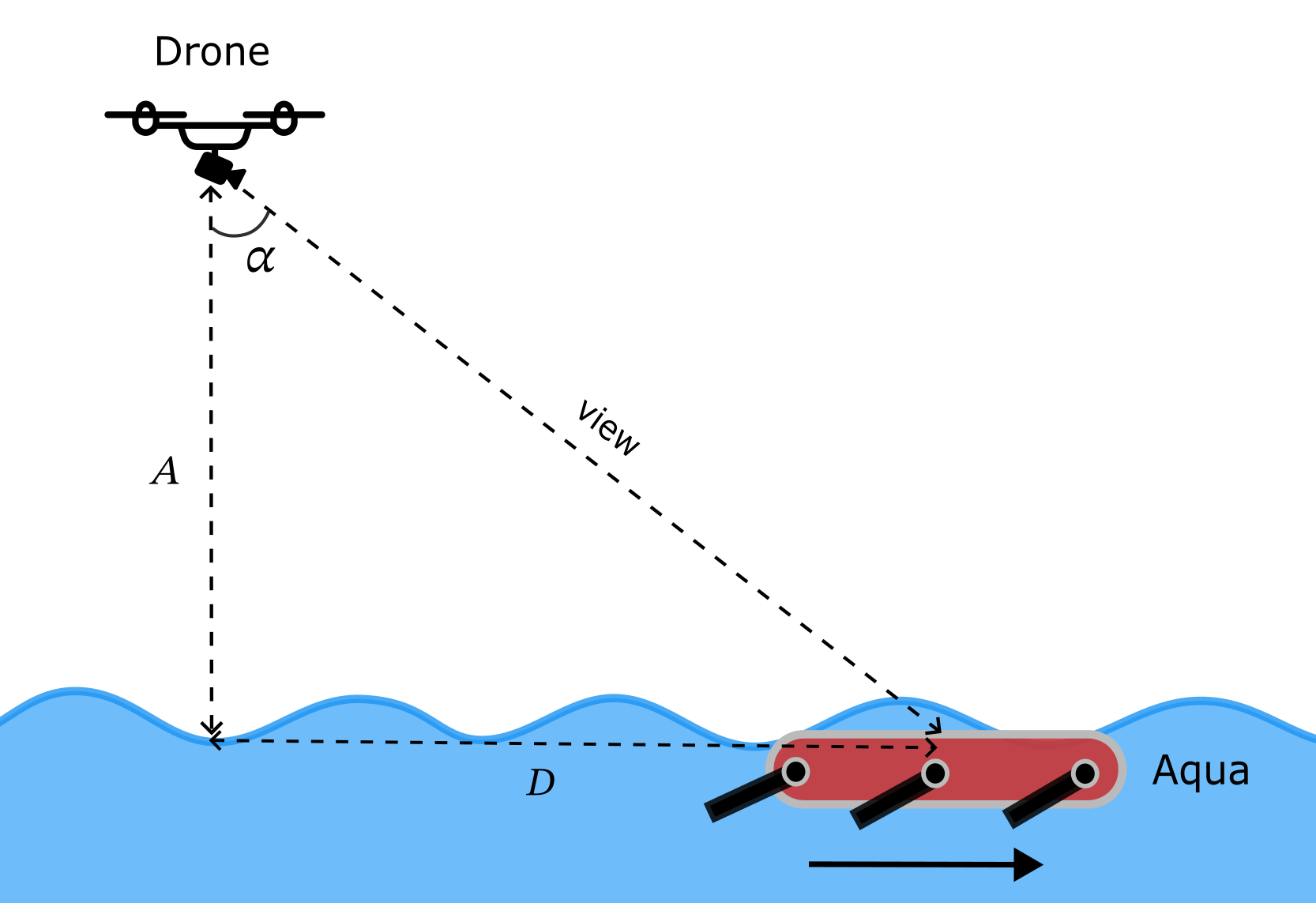} 
    \caption{Example of functional algorithm with a water robot and a drone.}
    \label{cartoon}
\end{figure}

In this paper, we adapt an existing aerial drone-based localization method \cite{drone_loc} to track and localize marine robots. This paper describes a preliminary exploration into a simple cost-effective, efficient, and scalable solution to the localization problem. Aerial drones equipped with geopositioning systems have become affordable and ubiquitous. We explore using such a drone to observe and localize a marine vehicle when it is sufficiently close to the surface. The algorithmic framework to do this, and the associated proof-of-concept, are the core contributions of this paper. Our proposed algorithm consists of three main components: data acquisition, visual localization, and estimation. The first component, data acquisition, is responsible for storing images of the submersible water robot when it is near the surface, along with information such as camera angles and the drone's compass direction. The visual localization portion of the algorithm focuses on robot detection and tracking. Lastly, the estimation section computes the GNSS coordinates of the robot based on drone observations and location. The robot used for the sake of these experiments is the Aqua2~\cite{dudek2007aqua} from Independent Robotics. The code we used to train and demonstrate our work is available at \url{https://github.com/stevvwen/aerial_gnss}.

\section{Background}
Accurate localization for marine robots is crucial for numerous applications, including ocean exploration \cite{ocean_exploration}, environmental monitoring \cite{env_surveillance}, underwater infrastructure inspection \cite{inf_inspect}, marine research \cite{marine_research}, and search and rescue operations \cite{search_rescue}. 

There are a range of different methods that can be used for underwater localization, each with its own tradeoff between cost, complexity, robustness, and accuracy. Most approaches rely on inertial sensors combined with Doppler Velocity Loggers (DVLs), which measure the robot's velocity relative to the surrounding water \cite{turbulent_ocean}, \cite{artic_dvl}. However, this navigation technique accumulates errors over time, resulting in positional drift \cite{dvl_error}. To address these limitations, acoustic-based techniques, notably Long Baseline (LBL) and Ultra Short Baseline (USBL) methods, have been used historically, where localization is achieved through the triangulation of acoustic signals from beacons or transponders \cite{lbl3}, \cite{lbl2}, \cite{lbl1}. Recent advancements, such as the 3D-BLUE system \cite{3d_blue}, employ piezo-electric backscatter technology to achieve accurate 3D localization in shallow underwater environments utilizing a single anchor to enable simultaneous localization of multiple underwater robots. 
Other research has shifted towards minimizing infrastructure through dynamic multi-agent systems and Simultaneous Localization and Mapping (SLAM) \cite{slam}. Various SLAM methodologies have been employed underwater, including Extended Kalman Filter (EKF) SLAM \cite{slam_fail1}, Sparse Extended Information Filter (SEIF) \cite{seif}, and FastSLAM \cite{fast}. However, these approaches often suffer from increasing computational complexity and reduced fidelity in environmental representation \cite{slam_fail1}, \cite{slam_fail2}. 

Given the limitations of traditional underwater localization methods, there remains a critical need for more efficient and scalable solutions. Recent innovations have begun leveraging hybrid cross-domain localization strategies. A hybrid localization framework explored in \cite{cross_view} correlates optical aerial images with acoustic underwater imagery to improve the localization of underwater vehicles, effectively addressing the challenge of aligning disparate modalities through domain adaptation and feature fusion. Another relevant approach involves cooperative bearing-only localization underwater, demonstrated in \cite{bearings}, by extending traditional visual bearing methods to underwater robots while integrating inertial, magnetic, and depth sensors. 

A cooperative localization framework for exploring underwater terrains is presented in \cite{multi_robots}, in which one robot is mainly responsible for data collection, while another provides assistance and localization support for the first one. This method enables robots to dynamically share mapping information and exploration targets, reducing mission time and enhancing operational efficiency.

To mitigate these issues, underwater robots need to surface periodically for positional recalibration. The availability of inexpensive GNSS modules in drones presents an effective solution for estimating the position of a surfaced marine robot.

The integration of drone-based GNSS localization has been previously explored, as demonstrated by prior studies such as \cite{drone_loc}, which utilize an Unmanned Aerial Vehicle (UAV) equipped with the You Only Look Once (YOLO) algorithm to detect objects from an image stream within a Robot Operating System (ROS) bag. The detected object's location is then determined using the UAV's own positional data. Additionally, the Sunflower system \cite{sunflower} employs a laser-based sensing mechanism combined with a queen-worker communication architecture between aerial drones and underwater vehicles, illustrating the substantial potential of drones in enhancing hybrid localization frameworks.

\section{Proposed Algorithm}
The proposed method to see and estimate the location of the marine robot comprises of 3 components. These are Data Acquisition, Visual Localization, and Estimation. 





\subsection{Data Acquisition}

A visual representation of the data acquisition can be seen in Figure~\ref{cartoon}. There is a submersible robot, in this case, an Aqua2, which is near the surface. The drone's camera has the marine robot in its field of view, and this image is temporarily stored. For the algorithm to work the drone would need to know its present height relative to the ocean, its GNSS coordinates, the angle of the camera, the pitch of the drone, the direction of the drone relative to the north, the focal length of the camera given the zoom, the sensor width of the camera, and finally the width and height of the image in pixels. 

This comprises all of the data required to be collected to estimate the position of a robot near the surface of water. The exact equation will also require knowledge of the position of the center of the robot relative to the image. This can be done using a pre-trained vision model such as YOLO.

\subsection{Data Augumentation}

Image augmentation is an effective technique for enhancing model performance and robustness \citep{perez2017effectiveness}, particularly when working with datasets of limited sizes \citep{1227801}. Given the challenges posed by significant variations in lighting conditions and image distortions, as well as the difficulty in obtaining abundant experimental data, we adopt this approach in our model.

Common data augmentation operations, such as rotation, flipping, cropping, padding, and affine transformations, are applied to the original images. Given the significant movement in our images, we apply motion blur. We also use the glass blur to simulate the light distortions caused by seawater. Since our real-life experiments are conducted at various times of the day, we adjust the color temperatures and brightness levels to account for changing lighting conditions. To ensure quality, we filter out any augmented images in which the center of the bounding box fell outside the image boundaries. All operations are independently sampled with a probability in random order. Due to the limitation that our vision model only supports square images, we crop the images to 1280x1280. Samples of the augmented data are shown in Figure~\ref{aug}.

\begin{figure}[h]
    \centering
    \includegraphics[width=0.3\textwidth]{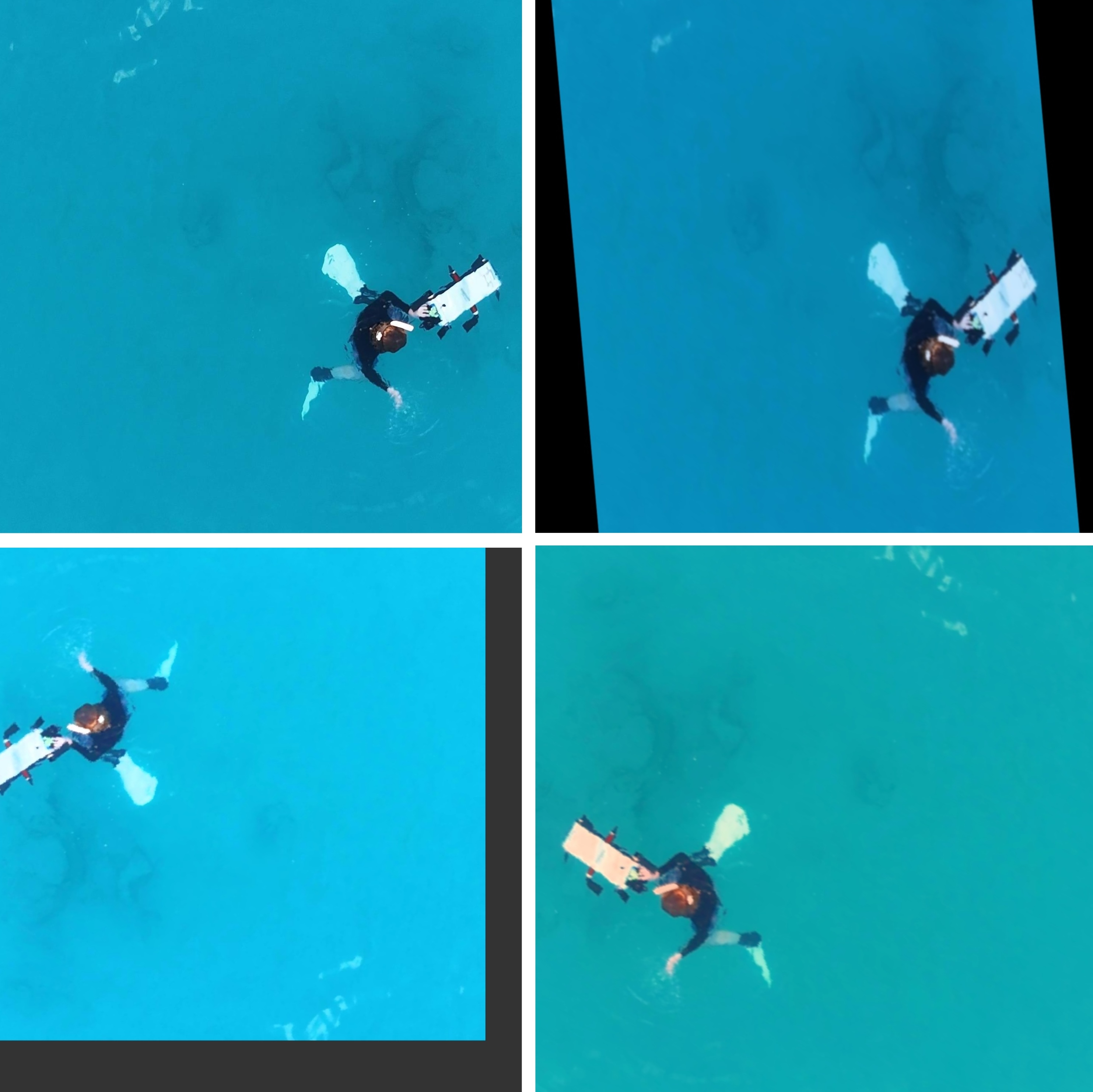} 
    \caption{Example of augmented images; the raw image is shown on the top left.}
    \label{aug}
\end{figure}

\subsection{Vision-based Localization via an Aerial Observer}

We approach this water marine robot localization task using computer vision, framing it as an object detection and tracking problem. This aligns closely with underwater tracking, where previous studies have highlighted the need for lightweight, powerful vision models capable of near real-time performance~\citep{8206280}. To meet these demanding requirements, we employ the convolution-based vision model called YOLO version 11~\citep{khanam2024yolov11}, which provides a good balance between computational efficiency and accuracy, making it well-suited for detecting and tracking objects in complex oceanic environments.

Our study places a strong emphasis on multi-robot localization while ensuring that the vision model performs effectively in both single- and multi-robot scenarios. To achieve this, we adopt a curriculum learning strategy to enhance performance progressively. The training process is divided into two distinct phases, leveraging a pre-trained YOLO model as the base. In the first phase, the model is trained for 100 epochs on the simpler task of single marine robot detection, allowing it to learn the fundamental features of the marine robot. In the second phase, the model undergoes an additional 100 epochs of training on the more complex task of detecting multiple robots simultaneously. This stepwise approach enables the model to build upon its initial knowledge, improving its ability to handle challenging underwater scenarios.

The vision model training is conducted using the default Ultralytics framework on a system equipped with an Nvidia RTX 4090 GPU (24 GB of GPU memory) and an AMD Ryzen 9 9950X 16-Core Processor (32 threads). The model processed images downsample to a 1280×1280 resolution, optimizing computational efficiency while ensuring high accuracy in underwater robot localization.

\subsection{Estimation}

This section details the mathematical formulation for computing the GNSS coordinates of an object observed near the water's surface by a drone with known GNSS coordinates. The derivation consists of two key steps: determining the estimated position of the object relative to the center of the image and computing its offset using camera parameters.

\subsubsection{Estimating the Position of the Object Relative to the Image Center}

\begin{figure}[h]
    \centering
    \includegraphics[width=0.4\textwidth]{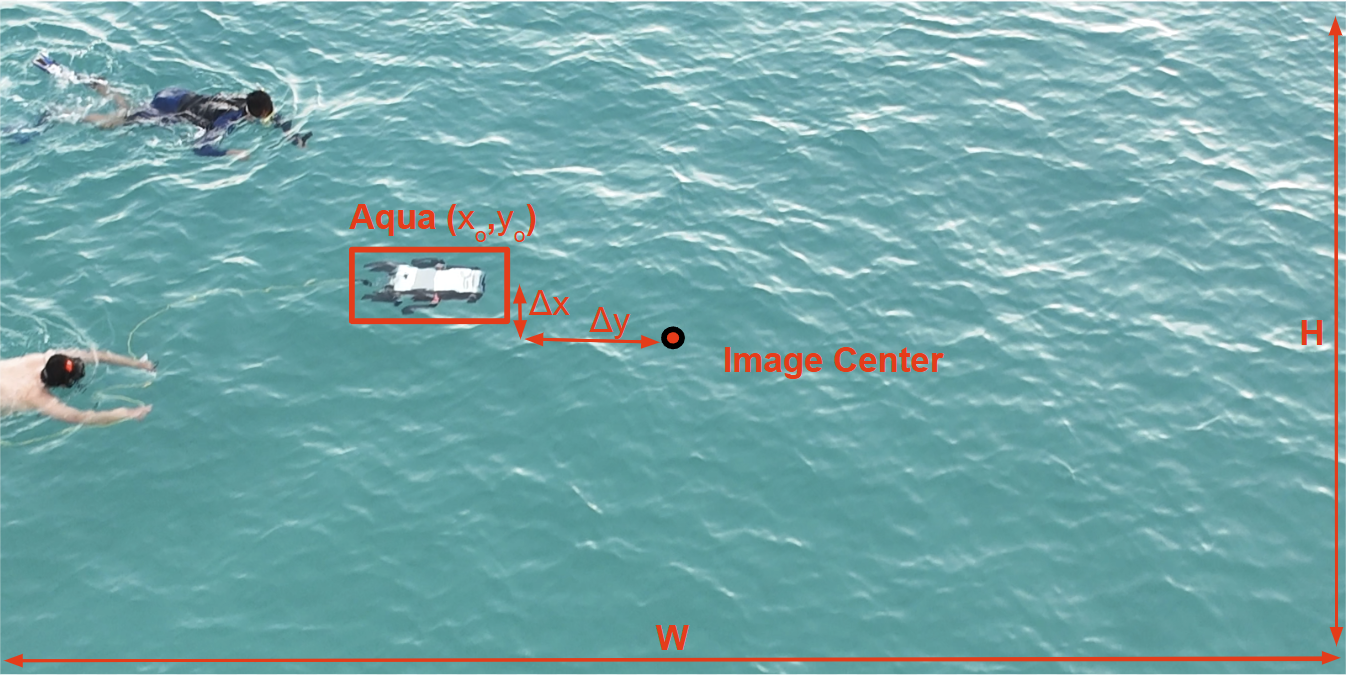} 
    \caption{Sample frame showing the offset of the marine robot (Aqua2) from the center of the image.}
    \label{offset}
\end{figure}

Given an image captured by a drone, the object might sometimes be located at the center of the image, in which case no further adjustment is required. However, as shown in Figure~\ref{offset}, the robot of interest may be offset from the center. To address this, we first calculate the angular displacement of the object relative to the image center. Suppose the image dimensions are \(W \times H\), where \(W\) is the width and \(H\) is the height in pixels. The object’s pixel coordinates \((x_o, y_o)\) are obtained using detection methods such as YOLO, as described earlier. For the simplicity of computation, we assumed the camera has yaw and roll angles to be zero.

The displacement of the object from the image center is computed as:
\begin{equation}
\Delta x = x_o - \frac{W}{2}, \quad \Delta y = y_o - \frac{H}{2}.
\end{equation}

Using the focal length of the camera $f$ and optics sensor dimensions $(S_x, S_y)$, we determine the field of view (FoV) in degrees along the horizontal and vertical axes:
\begin{equation}
\text{FoV}_x = 2 \tan^{-1}\left(\frac{S_x}{2f}\right), \quad \text{FoV}_y = 2 \tan^{-1}\left(\frac{S_y}{2f}\right).
\end{equation}

The angular displacement of the object from the center is then computed as:
\begin{equation}
\begin{split}
\theta_x &= \tan^{-1}\left(\left(\frac{2\Delta x}{W}\right) \tan \left(\frac{\text{FoV}_x}{2}\right)\right),\\[1ex]
\theta_y &= \tan^{-1}\left(\left(\frac{2\Delta y}{H}\right) \tan \left(\frac{\text{FoV}_y}{2}\right)\right)
\end{split}
\end{equation}

\begin{figure}[htbp]
  \centering
  \tdplotsetmaincoords{70}{120}

{\scriptsize
\begin{tikzpicture}[tdplot_main_coords, scale=2]

  \def\R{2}
  \def\H{1}
  \def\angle{20}

  \pgfmathsetmacro{\X}{\R*cos(\angle)}
  \pgfmathsetmacro{\Y}{\R*sin(\angle)}
  \coordinate (O) at (0,0,0);

  \pgfmathsetmacro{\tanalpha}{\H/\R}

  \coordinate (T) at (0, 0, \H);
  \coordinate (L) at (0.75*\R,\R,0);
  \coordinate (R) at (0,\R,0);

  \pgfmathsetmacro{\ax}{0.25*\R/tan(\angle)}

  \coordinate (C) at (0, 0.25*\R, 0.75*\H);
  \coordinate (A) at (0.1875*\R, 0.25*\R, 0.75 *\H);

  \draw[dashed] (0:\R) arc (0:360:\R);

  \draw[thick] (T) -- (L);       
  \draw[thick] (T) -- (R);  

    \filldraw[fill=gray!45, draw=blue!70, thick, opacity=0.5]
     ($ (-0.4, 0.1*\H, 0.1*\R)+ (C)$) -- ($ (0.4, 0.1*\H, 0.1*\R)+ (C)$) -- ($ (0.4, -0.1*\H, -0.1*\R)+ (C)$) -- ($ (-0.4, -0.1*\H, -0.1*\R)+ (C)$)  -- cycle;

     \draw[dashed, blue] ($ (-0.4,0, 0)+ (C)$) -- ($ (0.4,0, 0)+ (C)$);

    

  \filldraw[red] (C) circle (0.4pt);
  \filldraw[red] (A) circle (0.4pt);

    \draw[dashed, gray] (T) -- ($ (-0.4,  0.1*\H,  0.1*\R) + (C) $);
    \draw[dashed, gray] (T) -- ($ ( 0.4,  0.1*\H,  0.1*\R) + (C) $);
    \draw[dashed, gray] (T) -- ($ ( 0.4, -0.1*\H, -0.1*\R) + (C) $);
    \draw[dashed, gray] (T) -- ($ (-0.4, -0.1*\H, -0.1*\R) + (C) $);

  \draw[dashed] (0,0,0) -- (T) node[pos=0.35, left] {$A$};

  \draw[dashed] (0, 0, 0) -- (L) node[midway, below left, xshift=2pt, yshift=1pt] {$D_r$};
  \draw[dashed] (0, 0, 0) -- (R) node[midway, above right, xshift=2pt, yshift=1pt] {$D$};

  \draw[thick] (L) -- (R) node[midway, below right] {$D_x$};

  \filldraw[red] (T) circle (1pt) node[above=4pt, text=blue] {\small \textbf{drone}};
  \filldraw[red] (L) circle (1pt) node[below=4pt, text=blue] {\small \textbf{Robot}};

  \pic [draw, angle radius=15pt, angle eccentricity=1.3, "$\scriptstyle \theta_{x}$"] {angle = L--T--R};
  
  \path (0,0,0) -- (L) coordinate[pos=0.2] (LO);
  \path (0,0,0) -- (R) coordinate[pos=0.2] (RO);

  \path (0, 0, 0) -- (-2, 1, 0) coordinate[pos=0.2] (NO);

  \pic [draw, angle radius=20pt, angle eccentricity=1.4, "\scriptsize $\beta$"] {angle = LO--O--RO};

  \draw[->, dashed, brown](0, 0, 0) -- (-2, 1, 0) node [pos=0.8, left] {N};

  \pic [draw, brown, angle radius=15pt, angle eccentricity=1.5, "\scriptsize $\psi$"] {angle = RO--O--NO};



    
    

\end{tikzpicture}
}
  \caption{Demonstration of the 3D geometry}
  \label{fig:my-drawing}
\end{figure}
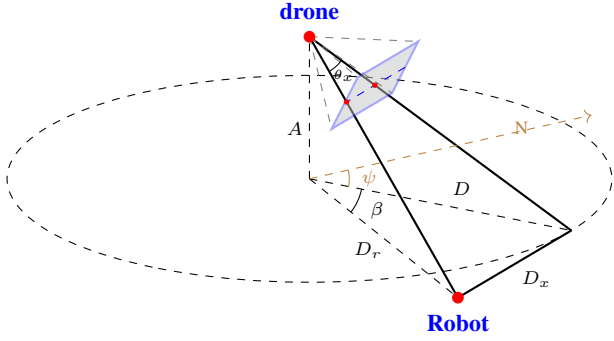

\subsubsection{Computing the Position}
With the angular displacement determined, the ground distance from the drone to the object can be computed. Given that the drone is at an altitude \(A\) and \(\alpha\) represents the combined downward angle of the camera and its pitch with the vertical axis, the ground distance \(D\) is given by:
\begin{equation}
    D = A \tan(\alpha + \theta_y).
\end{equation}

The lateral shift in the east-west direction due to the horizontal angle $\theta_x$ is given by:
\begin{equation}
D_x = \sqrt{A^2+D^2}\tan(\theta_x).
\end{equation}


The angular displacement of the robot with respect to the drone $\beta$ is given by:

\begin{equation}
    \beta= \tan^{-1} ( \frac{D_x}{D})
\end{equation}

The ground displacement of the robot which appears as a horizontal shift on the image plane $D_r$ is computed using trigonometry:

\begin{equation}
    D_r= \frac{\sqrt{\sin^2(\theta_x)A^2+ D^2}}{\cos(\theta_x)}
\end{equation}

The 3D demonstration of the model is shown in Figure~\ref{fig:my-drawing}.

Next, we resolve the displacement in terms of the drone's heading angle $\psi$ (relative to true north). The northward and eastward offsets $(\Delta N, \Delta E)$ of the object are:
\begin{equation}
\Delta N = D_r \cos(\psi+ \beta),
\end{equation}
\begin{equation}
\Delta E = D_r \sin(\psi+ \beta).
\end{equation}

To convert these distances to latitude and longitude shifts, we use the Earth's approximate conversion factors:
\begin{equation}
 \Delta \text{lat} = \frac{\Delta E}{\text{FPD Latitude}}.
\end{equation}

\begin{equation}
 \Delta \text{lon} = \frac{\Delta N}{\text{FPD Longitude}}.
\end{equation}

Where FPD is Feet Per Degree. Finally, the estimated GNSS coordinates of the object are obtained as:
\begin{equation}
\phi_{est} = \text{Latitude}{\text{-drone}} + \Delta \text{lat},
\end{equation}
\begin{equation}
\lambda_{est} = \text{Longitude}{\text{-drone}} + \Delta \text{lon}.
\end{equation}

This formulation provides a means to accurately estimate the position of an object near the surface of the water, observed by a drone, given its altitude, camera specifications, and heading information.

\subsection{Multi-Robot Localization}

Estimating the positions of multiple robots in a single frame fundamentally involves repeating the same estimation process for each detected instance. The underlying algorithm remains unchanged; it simply computes the position of every robot detected. The vision model, YOLO, in this case, must be trained to recognize and differentiate multiple instances of the same object or even various objects simultaneously \citep{redmon2018yolov3}. With robust detection in place, our proposed algorithm seamlessly performs multi-robot position estimation, ensuring accurate localization of each robot in the scene.

\begin{figure}[h]
    \centering
    \subfigure[Single Robot Sample 1: GNSS Position Estimation Error]{\includegraphics[width=0.24\textwidth]{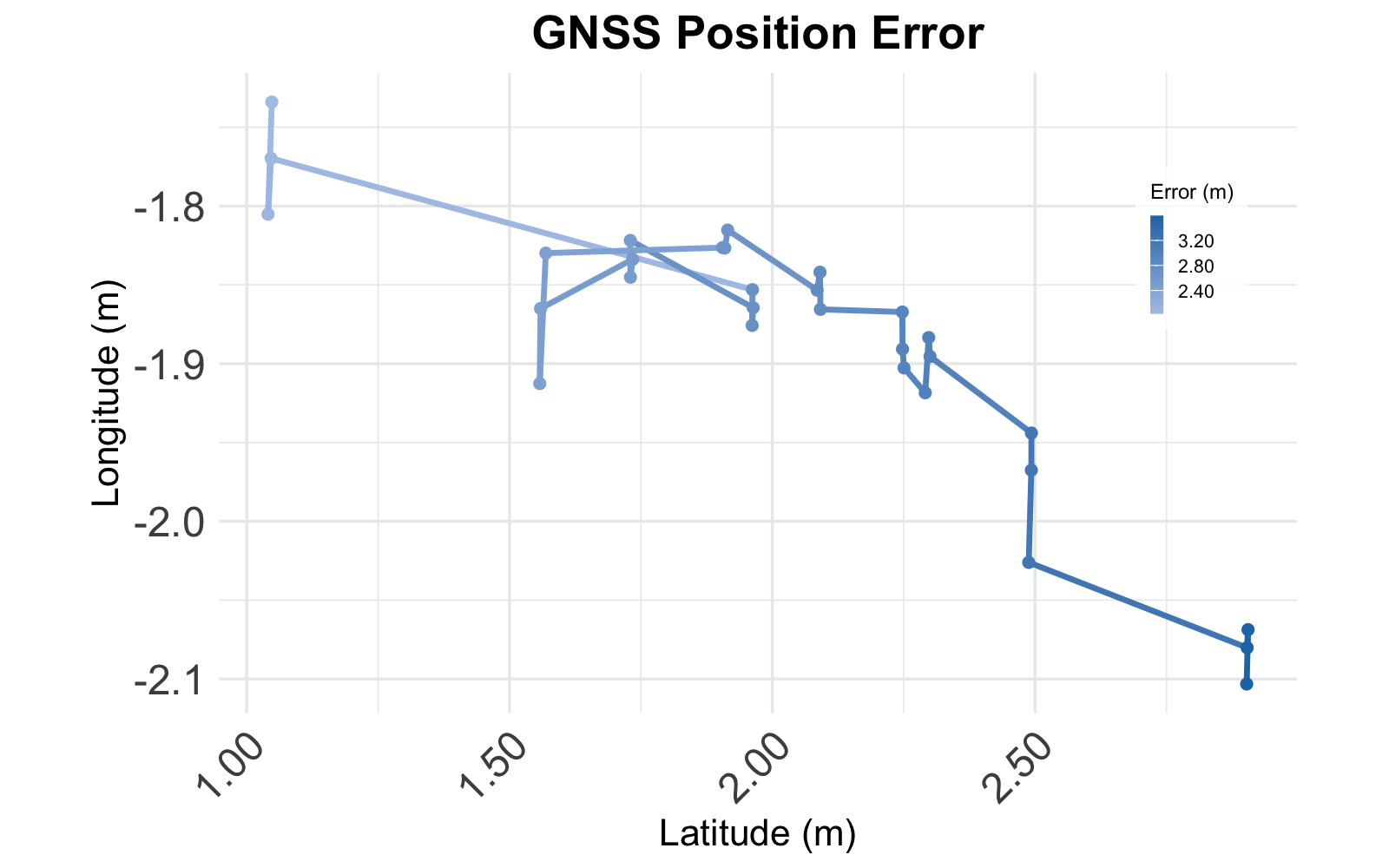}}
    \hfill
    \subfigure[Single Robot Sample 1: Localization Haversine error over frames]{\includegraphics[width=0.24\textwidth]{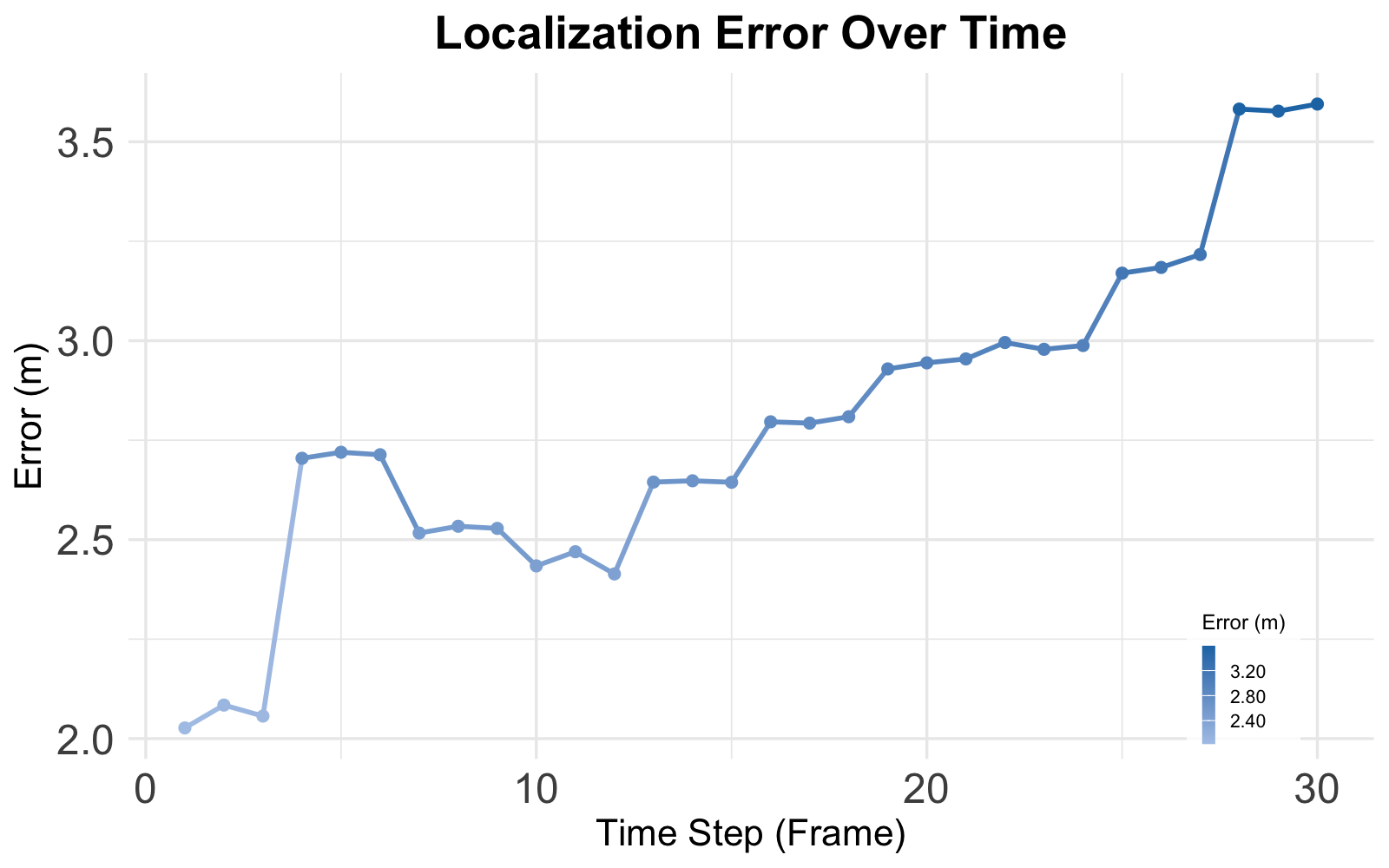}}
    \label{Aqua_single_yolo1}
\end{figure}

\begin{figure}[h]
    \centering
    \subfigure[Single Robot Sample 2: GNSS Position Estimation Error]{\includegraphics[width=0.24\textwidth]{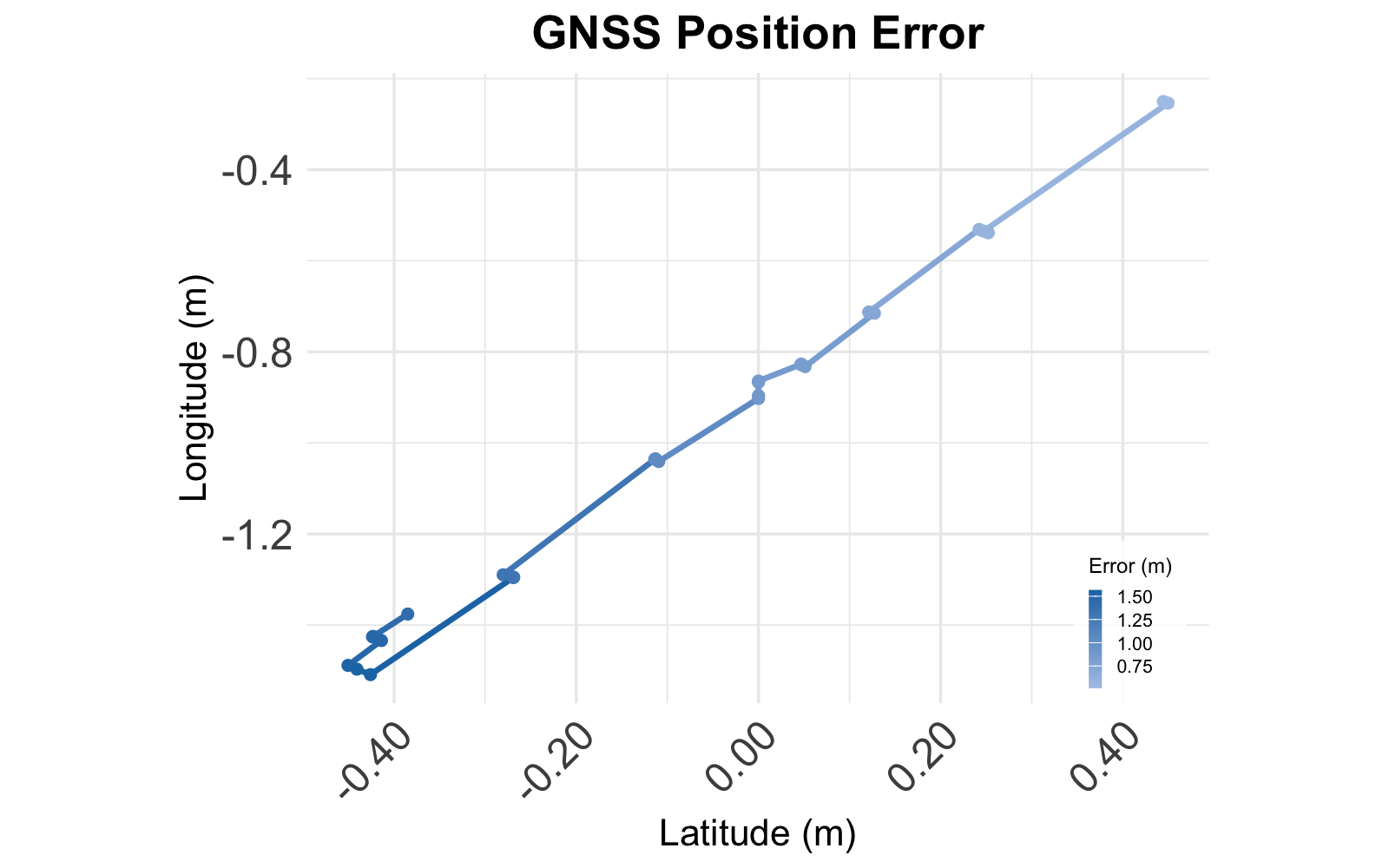}}
    \hfill
    \subfigure[Single Robot Sample 2: Localization Haversine error over frames]{\includegraphics[width=0.24\textwidth]{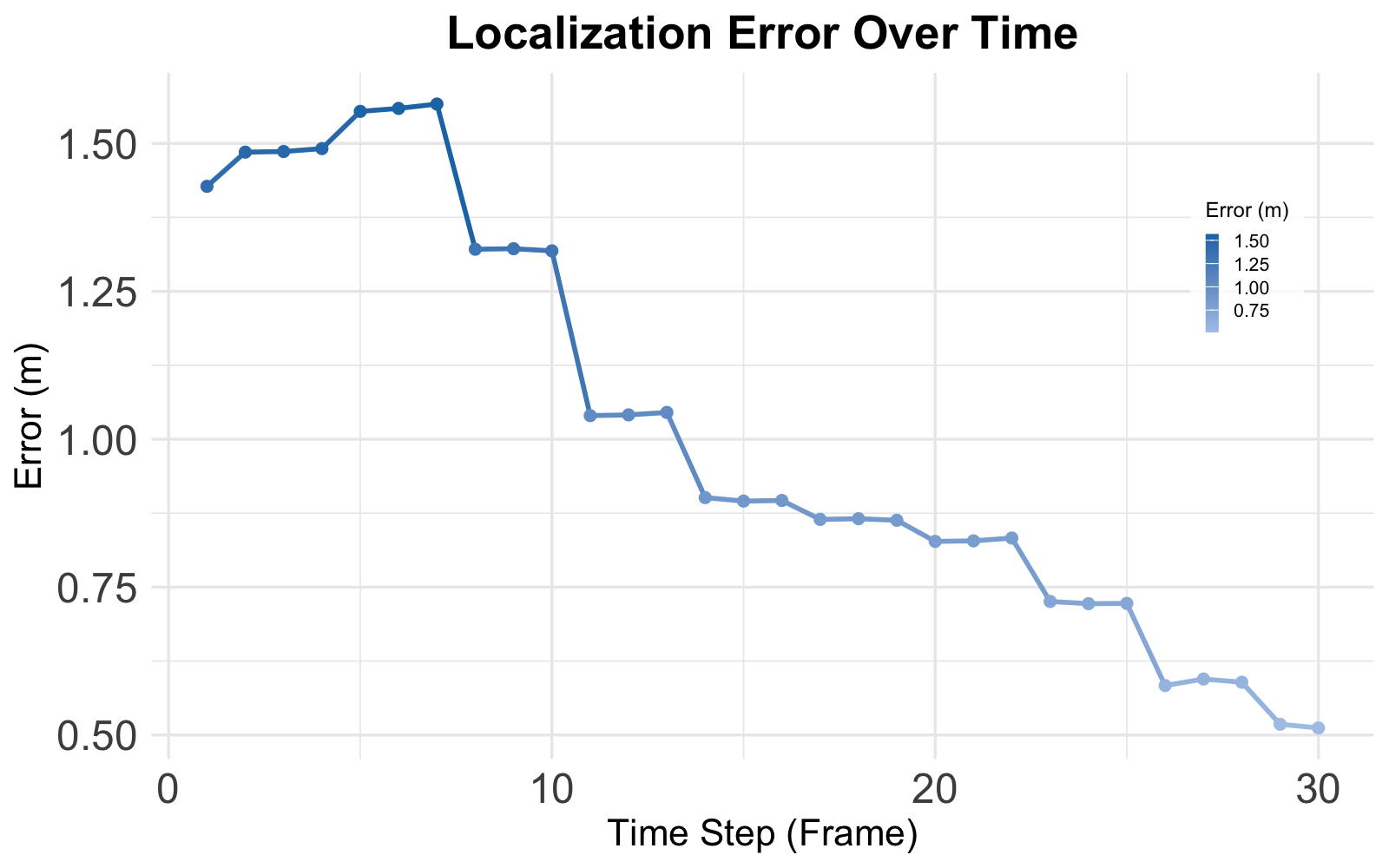}}
    \label{Aqua_single_yolo2}
\end{figure}

\begin{figure*}[ht]
    \centering
    \begin{minipage}{0.48\textwidth}
        \centering
        \includegraphics[width=\linewidth]{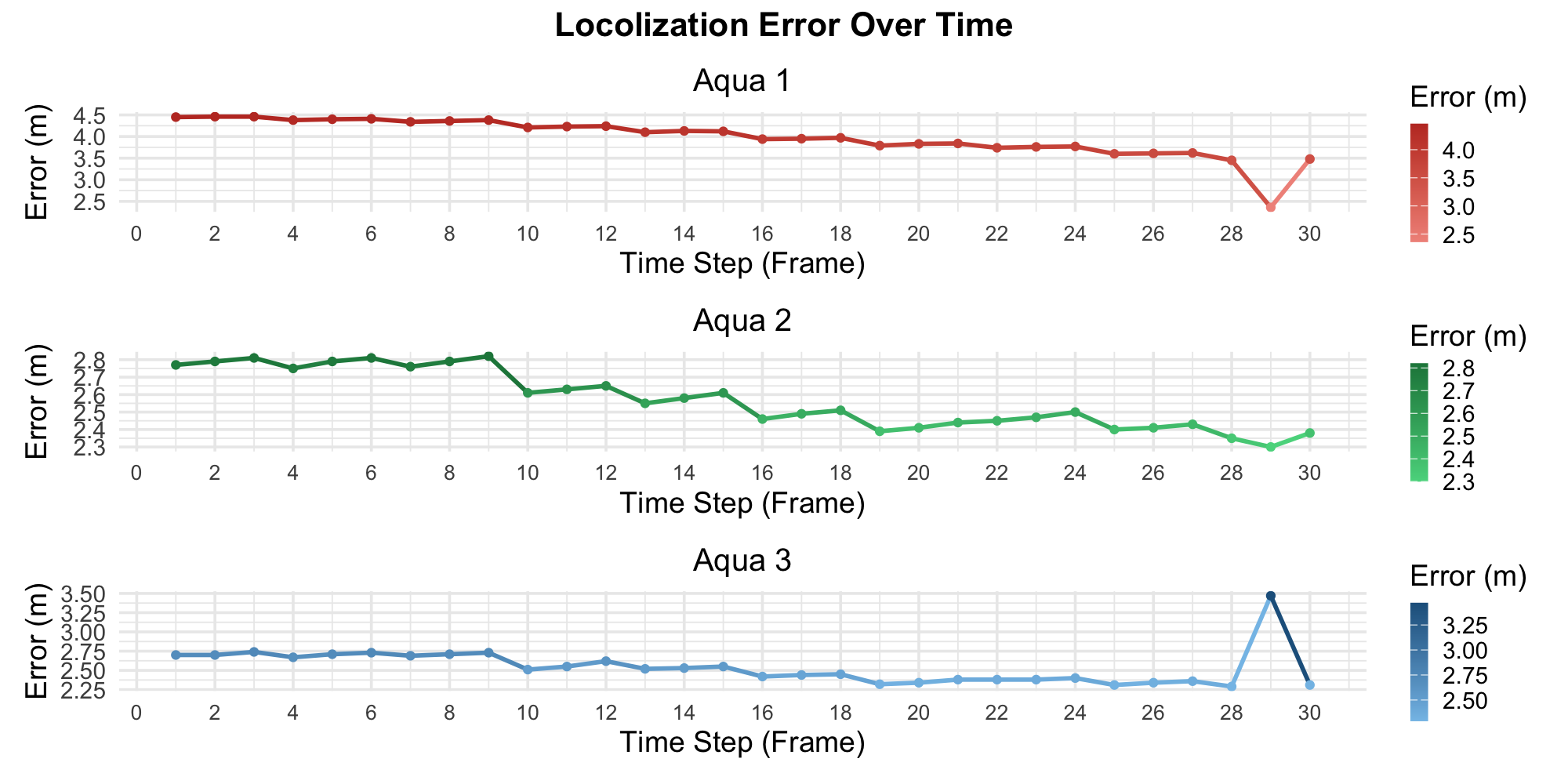}
        \caption{Multi Robot Sample 1: Localization Haversine error over frames}
        \label{fig:multi1_error}
    \end{minipage}\hfill
    \begin{minipage}{0.48\textwidth}
        \centering
        \includegraphics[width=\linewidth]{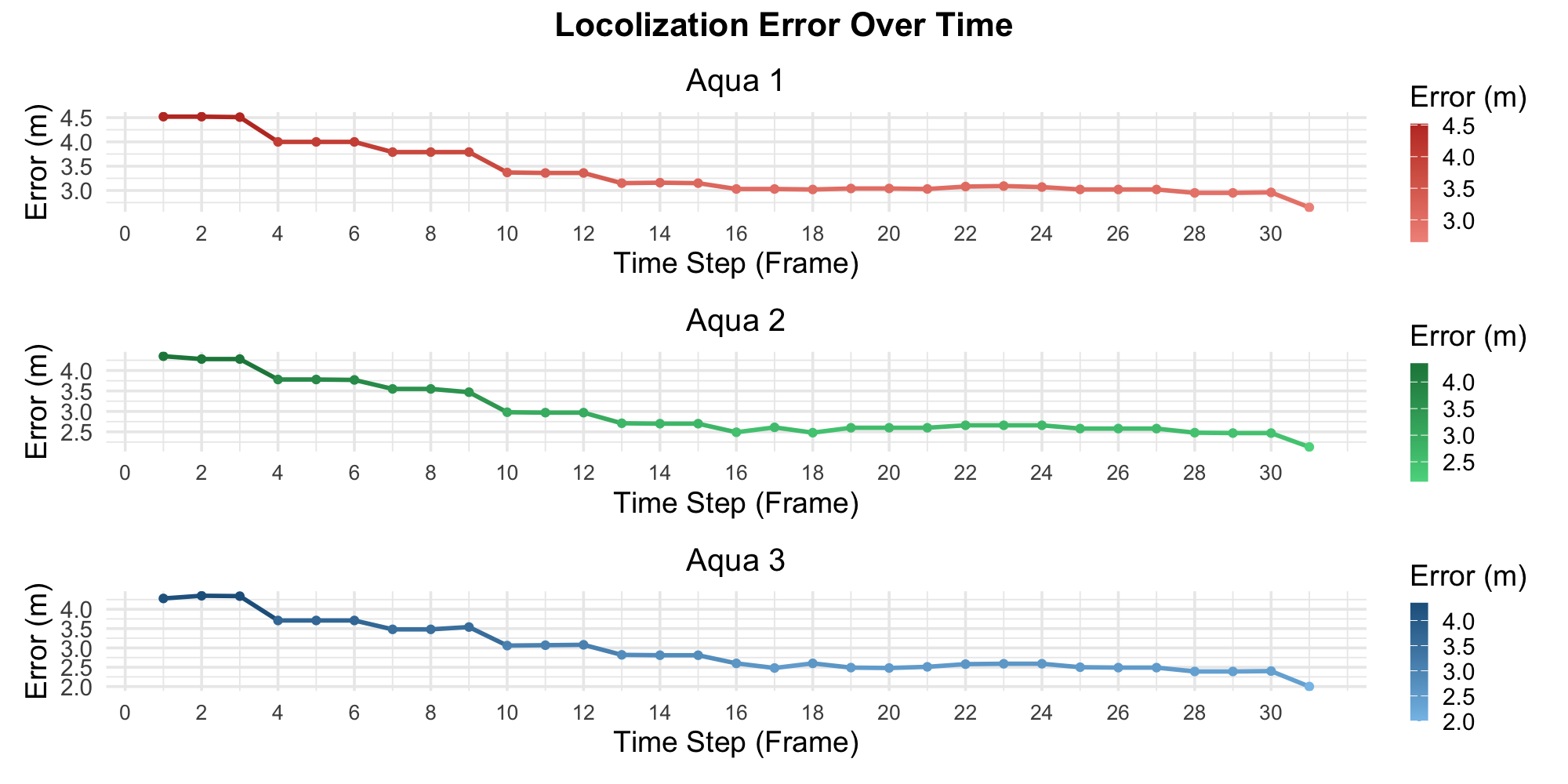}
        \caption{Multi Robot Sample 2: Localization Haversine error over frames}
        \label{fig:multi2_error}
    \end{minipage}
\end{figure*}

\begin{figure*}[ht]
    \centering
    \begin{minipage}{0.48\textwidth}
        \centering
        \includegraphics[width=1\linewidth]{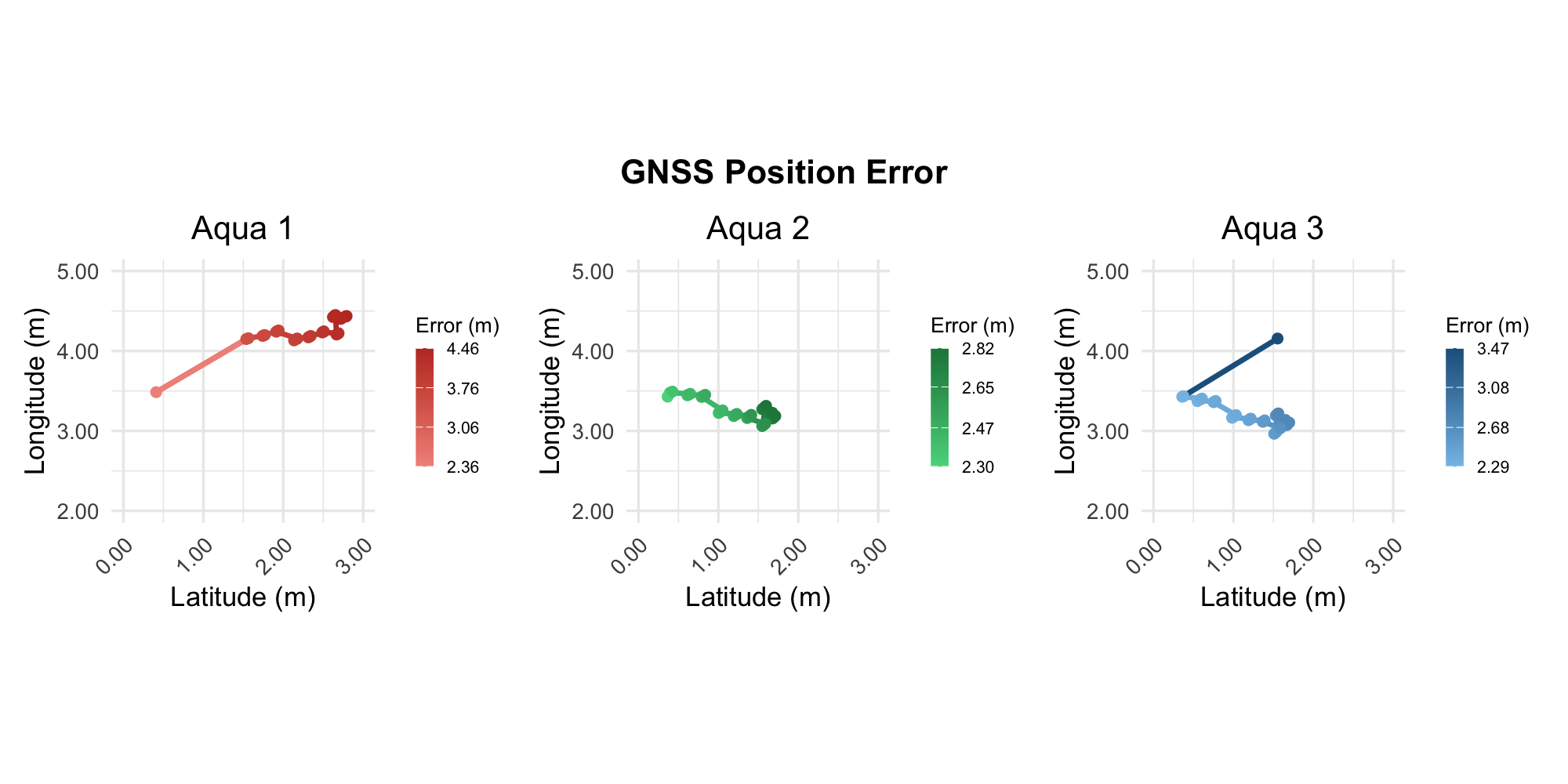}
        \caption{Multi Robot Sample 1: GNSS Position Estimation Error for multi Aqua}
        \label{fig:multi1_trajectory}
    \end{minipage}\hfill
    \begin{minipage}{0.48\textwidth}
        \centering
        \includegraphics[width=1\linewidth]{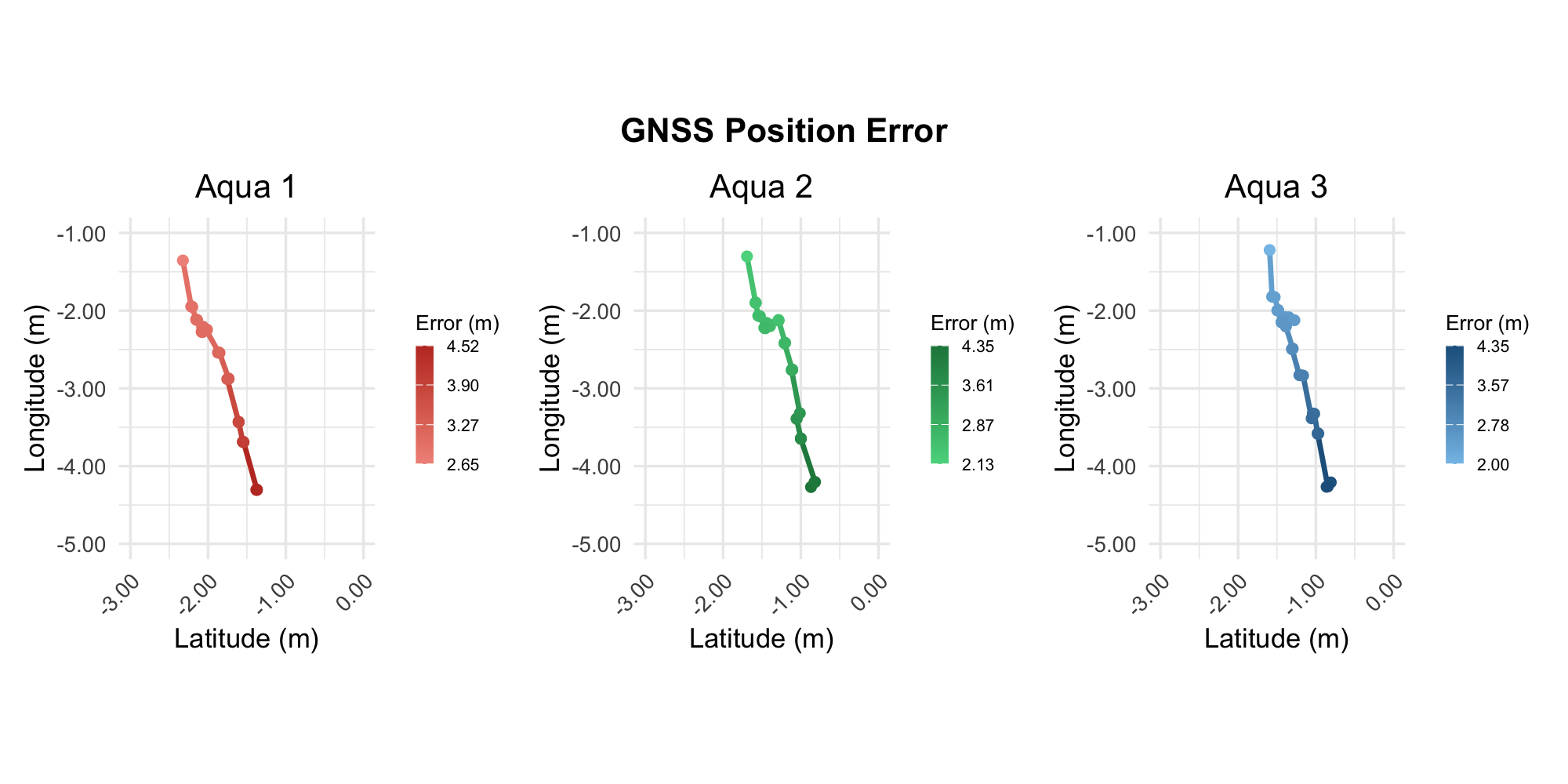}
        \caption{Multi Robot Sample 2: GNSS Position Estimation Error for multi Aqua}
        \label{fig:multi2_trajectory}
    \end{minipage}
\end{figure*}

\section{Experiments}

The experiments conducted consist of either one or three hexapod marine robot from the Aqua2 robot family being deployed and navigating underwater along a trajectory while remaining within 1 meter of the surface of the water. While the Aqua2 is navigating it is within the field of view of an overhead drone which records both video footage and telemetry data. A drawn-out example of this is seen in Figure~\ref{cartoon} with the image from the drone's point of view seen in Figure~\ref{offset}.

Before a robot begins moving or after it has stopped, the drone positions itself directly overhead to capture the robot’s precise GNSS coordinates. This is the true robot position. Once recorded, the drone moves back to a vantage point where all robots remain in view. A frame from the video, along with the collected data, is then used to estimate the GNSS positions. Upon computing this estimation, the coordinates are then compared to the previously recorded true coordinates using the Haversine formula~\cite{chopde2013landmark}:
\begin{equation}
\resizebox{0.85\columnwidth}{!}{$
\displaystyle
\Delta\sigma = 2 \arcsin\left(\sqrt{\sin^2\left(\frac{\Delta\phi}{2}\right) + \cos\phi_{est}\, \cos\phi_{tr}\, \sin^2\left(\frac{\Delta\lambda}{2}\right)}\right)
$}
\end{equation}

This manner of computing error provides a reliable measure of the distance in meters between the true and estimated positions.

As the drone was only able to either record the true GNSS positional data or collect the only data points have both sets of information (true and estimated data) are those when the aqua was being ready to deploy to finishing its mission. This will be improved upon in the future with the use of 2 drones.

\section{Results}

In this section, we present the results obtained from our field trials conducted using the described test methodology. Figures (a)–(d), 5-8 illustrate representative outcomes, demonstrating the effectiveness of our algorithm in accurately localizing submerged marine robots from drone imagery. Specifically, Figures (a)-(d) depict results from single-robot experiments, while Figures 5-8 illustrate results from trials involving three robots simultaneously. All localization estimations were computed at a frame rate of 30 frames per second, synchronized with the drone camera's capture rate.

The primary objective during these tests was to maintain localization errors within a few meters, as the drone's own positioning relies on GNSS localization, which typically has accuracy limitations of several meters~\cite{renfro2021analysis}.

\subsection{Single-Robot Localization}

The algorithm demonstrated consistent and accurate position estimation during single-robot trials. As shown in Figures (b) and (d), localization errors ranged from as low as $0.5$m to approximately $3.5$m under optimal conditions characterized by minimal sensor noise and accurate detection calibration. Further decomposition of the localization errors into longitude and latitude, shown in Figures (a) and (c), reveals consistent and stable estimation in both directions. Although some localization errors remain inevitable, the observed performance comfortably surpasses typical GNSS accuracy standards, validating the effectiveness of the proposed method in ideal scenarios.

\subsection{Multi-Robot Localization}

To assess the scalability of our algorithm, we further evaluated its performance in scenarios involving multiple robots simultaneously visible in the drone’s field of view. Figures~\ref{fig:multi1_error} and ~\ref{fig:multi2_error} illustrate the results from these multi-robot experiments, highlighting the algorithm’s capability to independently and accurately localize each robot with errors ranging between $2$m and $4.5$m.  breakdown of the localization errors along the longitude and latitude axes, shown in Figures ~\ref{fig:multi1_trajectory} and ~\ref{fig:multi2_trajectory}, indicates that the estimation remains stable across both directions, even in multi-robot scenarios. Despite the increased complexity introduced by multiple concurrent detections, the localization error remains acceptably low, demonstrating robust performance in more challenging conditions.

An additional observation from the trials is that localization errors tend to decrease when the drone is positioned closer to the robots. This relationship is anticipated, as primary sources of error—including sensor noise, detection inaccuracies, and uncertainties in altitude estimation—typically become more pronounced at greater distances. To provide further insight into these errors and their causes, we explore the main contributing factors in greater detail in the subsequent section.

\subsection{Sources of Error}
Despite the promising results, several primary sources of error were identified:
\begin{itemize}
    \item \textbf{Sensor Noise:} The localization computation depends on various sensor measurements (e.g., GNSS coordinates, altitude, camera angles). Inherent inaccuracies in these sensors contribute to the overall error.
    \item \textbf{Object Detection Uncertainty:} The position of each robot is determined by detecting its center using a fine-tuned YOLO model. While effective, slight inaccuracies in object detection can lead to deviations in the estimated position.
    \item \textbf{Tides and Currents:} Our localization approach relies heavily on the accurate measurement of the drone's altitude above sea level. However, this measurement can be distorted by changing tide levels and dynamic ocean currents. At different times of the day, the water level can vary significantly due to tidal cycles, leading to a discrepancy between the assumed and actual altitude of the target relative to the drone. Moreover, strong ocean currents may influence the robot's true altitude, introducing additional errors not accounted for in a static world model.
    \item \textbf{Misalignment Error:} To approximate the GNSS position of the robot, we use a second drone hovering above it. However, due to water currents and air turbulence, both the robot and the drone experience slight movements relative to one another, resulting in potential misalignment and positional error.

\end{itemize}

Overall, the experimental results confirm that the proposed method is capable of providing robust and accurate localization for aquatic robots near the surface of the water. The observed errors align with our analysis, and they highlight potential areas for further refinement—such as improved sensor calibration and enhanced detection precision—to further boost performance.

\subsection{Hardware Testing}

To show our approach can be deployed effectively in real-time, we conduct inference deployment on an NVIDIA Jetson AGX Xavier (32GB). This compute unit, suitable for use on a drone~\cite{rojas2021board}, is tested to determine how efficiently it could perform positional estimation using the proposed pipeline.  

In a scenario where only one marine robot needed to be detected, the Xavier processes the full pipeline approximately every 0.22 seconds, based on an average of over 3,000 frame estimations. This corresponds to a position update rate of around 4~Hz, meaning a marine robot could request and receive its estimated coordinates from the drone at this frequency before submerging again.  

More interestingly, performance is assessed in a multi-target setting. When detecting three marine robots simultaneously, the Xavier processes the full pipeline in approximately 0.26 seconds per frame (again averaged over 3,000 estimations). This demonstrates that multiple robots can be localized concurrently without significant computational overhead.  

These results suggest that Xavier's full processing capacity is not strictly necessary for this task alone when deployed on a drone, potentially allowing for resource allocation to additional operations or the use of a lower-power processor.  

\section{Conclusion}

In this paper, we examine a simple and efficient method for localizing near-surface marine vehicles using a drone equipped with a GNSS system and a learned appearance model based on the YOLO architecture. This approach significantly reduces localization costs, as a single drone can simultaneously determine the positions of multiple marine vehicles, providing an economic advantage compared to integrating specialized GNSS systems directly into each marine robot. For perspective, the cost of one specialized marine localization system is equivalent to approximately four drones similar to those utilized in our experiments.

By applying various data augmentation techniques during training, we enhanced the robustness and reliability of the YOLO model, enabling accurate localization results even with a relatively small dataset. Experimental results conducted under realistic maritime conditions demonstrate that our approach effectively handles both single-robot and multi-robot scenarios, underscoring its scalability and practicality for real-world marine robotics missions.

Future work will focus on further validating the precision of our model by introducing a second drone that tracks marine vehicles to independently verify their global positions. Additionally, exploring the potential improvements in accuracy achievable through collaborative position estimation using multiple drones remains a promising area of investigation.

\section{Acknowledgments}
We gratefully acknowledge the participants of the 2025 Annual Marine Robotics Workshop and Field Robotics Trials in Barbados for their invaluable support during our experiments. We also thank the students and broader community of the McGill Mobile Robotics Lab (MRL) and the Centre for Intelligent Machines (CIM) for their ongoing contributions. Special thanks go to Marios Xanthidis for his generous time and effort in helping us conduct the experiments. Finally, we extend our sincere appreciation to everyone at Independent Robotics for their crucial assistance in deploying the robots in both Montreal and Barbados.


\bibliography{biblio}

@article{renfro2021analysis,
  title={An analysis of global positioning system standard positioning service performance for 2020},
  author={Renfro, Brent A and Stein, Miquela and Reed, Emery B and Villalba, Eduardo J},
  journal={Space and Geophysics Laboratory Applied Research Laboratories The University of Texas at Austin. Available online: https://www. gps. gov/systems/gps/performance/(accessed on 1 September 2023)},
  year={2021}
}

@article{chopde2013landmark,
  title={Landmark based shortest path detection by using A* and Haversine formula},
  author={Chopde, Nitin R and Nichat, Mangesh},
  journal={International Journal of Innovative Research in Computer and Communication Engineering},
  volume={1},
  number={2},
  pages={298--302},
  year={2013}
}

@article{dudek2007aqua,
  title={Aqua: An amphibious autonomous robot},
  author={Dudek, Gregory and Giguere, Philippe and Prahacs, Chris and Saunderson, Shane and Sattar, Junaed and Torres-Mendez, Luz-Abril and Jenkin, Michael and German, Andrew and Hogue, Andrew and Ripsman, Arlene and others},
  journal={Computer},
  volume={40},
  number={1},
  pages={46--53},
  year={2007},
  publisher={IEEE}
}

@article{rojas2021board,
  title={On-board processing for autonomous drone racing: An overview},
  author={Rojas-Perez, L Oyuki and Mart{\'\i}nez-Carranza, Jos{\'e}},
  journal={Integration},
  volume={80},
  pages={46--59},
  year={2021},
  publisher={Elsevier}
}

@article{perez2017effectiveness,
  title={The effectiveness of data augmentation in image classification using deep learning},
  author={Perez, Luis and Wang, Jason},
  journal={arXiv preprint arXiv:1712.04621},
  year={2017}
}

@article{artic_dvl,
  author={McEwen, R. and Thomas, H. and Weber, D. and Psota, F.},
  journal={IEEE Journal of Oceanic Engineering}, 
  title={Performance of an AUV navigation system at Arctic latitudes}, 
  year={2005},
  volume={30},
  number={2},
  pages={443-454},
  doi={10.1109/JOE.2004.838336}}

@INPROCEEDINGS{turbulent_ocean,
  author={Garau, B. and Alvarez, A. and Oliver, G.},
  booktitle={Proceedings 2006 IEEE International Conference on Robotics and Automation, 2006. ICRA 2006.}, 
  title={AUV navigation through turbulent ocean environments supported by onboard H-ADCP}, 
  year={2006},
  volume={},
  number={},
  pages={3556-3561},
  doi={10.1109/ROBOT.2006.1642245}}

@Article{dvl_error,
AUTHOR = {He, Bo and Zhang, Hongjin and Li, Chao and Zhang, Shujing and Liang, Yan and Yan, Tianhong},
TITLE = {Autonomous Navigation for Autonomous Underwater Vehicles Based on Information Filters and Active Sensing},
JOURNAL = {Sensors},
VOLUME = {11},
YEAR = {2011},
NUMBER = {11},
PAGES = {10958--10980},
URL = {https://www.mdpi.com/1424-8220/11/11/10958},
PubMedID = {22346682},
ISSN = {1424-8220},
DOI = {10.3390/s111110958}
}

@INPROCEEDINGS{lbl1,
  author={Ribas, David and Ridao, Pere and Mallios, Angelos and Palomeras, Narcís},
  booktitle={2012 IEEE International Conference on Robotics and Automation}, 
  title={Delayed state information filter for USBL-Aided AUV navigation}, 
  year={2012},
  volume={},
  number={},
  pages={4898-4903},
  doi={10.1109/ICRA.2012.6224989}}

@ARTICLE{lbl2,
  author={Kussat, N.H. and Chadwell, C.D. and Zimmerman, R.},
  journal={IEEE Journal of Oceanic Engineering}, 
  title={Absolute positioning of an autonomous underwater vehicle using GPS and acoustic measurements}, 
  year={2005},
  volume={30},
  number={1},
  pages={153-164},
  doi={10.1109/JOE.2004.835249}}

@article{lbl3,
author = {B. Bingham and W. Seering},
title = {Hypothesis Grids: Improving Long Baseline Navigation for Autonomous Underwater Vehicles},
journal = {IEEE Journal of Oceanic Engineering},
year = {2006},
volume = {31},
publisher = {Institute of Electrical and Electronics Engineers (IEEE)},
month = {jan},
url = {https://doi.org/10.1109/joe.2006.872220},
number = {1},
pages = {209--218},
doi = {10.1109/joe.2006.872220}
}

@INPROCEEDINGS{1227801,
  author={Simard, P.Y. and Steinkraus, D. and Platt, J.C.},
  booktitle={Seventh International Conference on Document Analysis and Recognition, 2003. Proceedings.}, 
  title={Best practices for convolutional neural networks applied to visual document analysis}, 
  year={2003},
  volume={},
  number={},
  pages={958-963},
  doi={10.1109/ICDAR.2003.1227801}}

@ARTICLE{slam_fail1,
  author={Eustice, Ryan M. and Pizarro, Oscar and Singh, Hanumant},
  journal={IEEE Journal of Oceanic Engineering}, 
  title={Visually Augmented Navigation for Autonomous Underwater Vehicles}, 
  year={2008},
  volume={33},
  number={2},
  pages={103-122},
  doi={10.1109/JOE.2008.923547}}

@INPROCEEDINGS{slam_fail2,
  author={Newman, P. and Leonard, J.},
  booktitle={2003 IEEE International Conference on Robotics and Automation (Cat. No.03CH37422)}, 
  title={Pure range-only sub-sea SLAM}, 
  year={2003},
  volume={2},
  number={},
  pages={1921-1926 vol.2},
  doi={10.1109/ROBOT.2003.1241875}}

@INPROCEEDINGS{slam,
  author={Salvi, Joaquim and Petillo, Yvan and Thomas, Stephen and Aulinas, Josep},
  booktitle={OCEANS 2008}, 
  title={Visual SLAM for underwater vehicles using video velocity log and natural landmarks}, 
  year={2008},
  volume={},
  number={},
  pages={1-6},
  doi={10.1109/OCEANS.2008.5151887}}

@INPROCEEDINGS{drone_loc,
  author={Sanyal, Snehil and Bhushan, Shashank and Sivayazi, K},
  booktitle={2020 First International Conference on Power, Control and Computing Technologies (ICPC2T)}, 
  title={Detection and Location Estimation of Object in Unmanned Aerial Vehicle using Single Camera and GPS}, 
  year={2020},
  volume={},
  number={},
  pages={73-78},
  doi={10.1109/ICPC2T48082.2020.9071439}}

@INPROCEEDINGS{8206280,
  author={Shkurti, Florian and Chang, Wei-Di and Henderson, Peter and Islam, Md Jahidul and Higuera, Juan Camilo Gamboa and Li, Jimmy and Manderson, Travis and Xu, Anqi and Dudek, Gregory and Sattar, Junaed},
  booktitle={2017 IEEE/RSJ International Conference on Intelligent Robots and Systems (IROS)}, 
  title={Underwater multi-robot convoying using visual tracking by detection}, 
  year={2017},
  volume={},
  number={},
  pages={4189-4196},
  doi={10.1109/IROS.2017.8206280}}

@ARTICLE{seif,
  author={Eustice, Ryan M. and Singh, Hanumant and Leonard, John J.},
  journal={IEEE Transactions on Robotics}, 
  title={Exactly Sparse Delayed-State Filters for View-Based SLAM}, 
  year={2006},
  volume={22},
  number={6},
  pages={1100-1114},
  doi={10.1109/TRO.2006.886264}}

@article{fast,
author = {Stephen Barkby and Stefan B Williams and Oscar Pizarro and Michael V Jakuba},
title ={Bathymetric particle filter SLAM using trajectory maps},

journal = {The International Journal of Robotics Research},
volume = {31},
number = {12},
pages = {1409-1430},
year = {2012},
doi = {10.1177/0278364912459666},

URL = { https://doi.org/10.1177/0278364912459666
},
eprint = { https://doi.org/10.1177/0278364912459666
}
}

@article{khanam2024yolov11,
  title={Yolov11: An overview of the key architectural enhancements},
  author={Khanam, Rahima and Hussain, Muhammad},
  journal={arXiv preprint arXiv:2410.17725},
  year={2024}
}

@article{env_surveillance,
author = {Su, Xin and Ullah, Inam and Liu, Xiaofeng and Choi, Dongmin},
title = {A Review of Underwater Localization Techniques, Algorithms, and Challenges},
journal = {Journal of Sensors},
volume = {2020},
number = {1},
pages = {6403161},
doi = {https://doi.org/10.1155/2020/6403161},
url = {https://onlinelibrary.wiley.com/doi/abs/10.1155/2020/6403161},
eprint = {https://onlinelibrary.wiley.com/doi/pdf/10.1155/2020/6403161},
year = {2020}
}

@Article{ocean_exploration,
AUTHOR = {Barker, Laughlin D. L. and Jakuba, Michael V. and Bowen, Andrew D. and German, Christopher R. and Maksym, Ted and Mayer, Larry and Boetius, Antje and Dutrieux, Pierre and Whitcomb, Louis L.},
TITLE = {Scientific Challenges and Present Capabilities in Underwater Robotic Vehicle Design and Navigation for Oceanographic Exploration Under-Ice},
JOURNAL = {Remote Sensing},
VOLUME = {12},
YEAR = {2020},
NUMBER = {16},
ARTICLE-NUMBER = {2588},
URL = {https://www.mdpi.com/2072-4292/12/16/2588},
ISSN = {2072-4292},
DOI = {10.3390/rs12162588}
}

@article{inf_inspect,
  author    = {Athanasios C. Kapoutsis and Savvas A. Chatzichristofis and Lefteris Doitsidis},
  title     = {Real-time adaptive multi-robot exploration with application to underwater map construction},
  journal   = {Autonomous Robots},
  year      = {2016},
  volume    = {40},
  number    = {6},
  pages     = {987--1011},
  doi       = {10.1007/s10514-015-9510-8},
  url       = {https://link.springer.com/article/10.1007/s10514-015-9510-8},
  publisher = {Springer}
}

@InProceedings{marine_research,
author="Whitcomb, Louis
and Yoerger, Dana R.
and Singh, Hanumant
and Howland, Jonathan",
editor="Hollerbach, John M.
and Koditschek, Daniel E.",
title="Advances in Underwater Robot Vehicles for Deep Ocean Exploration: Navigation, Control, and Survey Operations",
booktitle="Robotics Research",
year="2000",
publisher="Springer London",
address="London",
pages="439--448",
isbn="978-1-4471-0765-1"
}

@article{search_rescue,
title = {Multi-Robot Exploration of Underwater Structures},
journal = {IFAC-PapersOnLine},
volume = {55},
number = {31},
pages = {395-400},
year = {2022},
note = {14th IFAC Conference on Control Applications in Marine Systems, Robotics, and Vehicles CAMS 2022},
issn = {2405-8963},
doi = {https://doi.org/10.1016/j.ifacol.2022.10.460},
url = {https://www.sciencedirect.com/science/article/pii/S240589632202506X},
author = {Marios Xanthidis and Bharat Joshi and Jason M. O'Kane and Ioannis Rekleitis}
}

@INPROCEEDINGS{3d_blue,
  author={Afzal, Sayed Saad and Chen, Weitung and Adib, Fadel},
  booktitle={2024 IEEE/RSJ International Conference on Intelligent Robots and Systems (IROS)}, 
  title={3D-BLUE: Backscatter Localization for Underwater Robotics}, 
  year={2024},
  volume={},
  number={},
  pages={11040-11047},
  doi={10.1109/IROS58592.2024.10801869}}

@ARTICLE{cross_view,
  author={Santos, Matheus M. Dos and De Giacomo, Giovanni G. and Drews-Jr, Paulo L. J. and Botelho, Silvia S. C.},
  journal={IEEE Robotics and Automation Letters}, 
  title={Cross-View and Cross-Domain Underwater Localization Based on Optical Aerial and Acoustic Underwater Images}, 
  year={2022},
  volume={7},
  number={2},
  pages={4969-4974},
  doi={10.1109/LRA.2022.3154482}}

@inproceedings{sunflower,
author = {Carver, Charles J. and Shao, Qijia and Lensgraf, Samuel and Sniffen, Amy and Perroni-Scharf, Maxine and Gallant, Hunter and Li, Alberto Quattrini and Zhou, Xia},
title = {Sunflower: locating underwater robots from the air},
year = {2022},
isbn = {9781450391856},
publisher = {Association for Computing Machinery},
address = {New York, NY, USA},
url = {https://doi.org/10.1145/3498361.3539773},
doi = {10.1145/3498361.3539773},
booktitle = {Proceedings of the 20th Annual International Conference on Mobile Systems, Applications and Services},
pages = {14–27},
numpages = {14},
location = {Portland, Oregon},
series = {MobiSys '22}
}

@article{redmon2018yolov3,
  title={Yolov3: An incremental improvement},
  author={Redmon, Joseph and Farhadi, Ali},
  journal={arXiv preprint arXiv:1804.02767},
  year={2018}
}

@article{multi_robots,
title = {Multi-Robot Exploration of Underwater Structures},
journal = {IFAC-PapersOnLine},
volume = {55},
number = {31},
pages = {395-400},
year = {2022},
note = {14th IFAC Conference on Control Applications in Marine Systems, Robotics, and Vehicles CAMS 2022},
issn = {2405-8963},
doi = {https://doi.org/10.1016/j.ifacol.2022.10.460},
url = {https://www.sciencedirect.com/science/article/pii/S240589632202506X},
author = {Marios Xanthidis and Bharat Joshi and Jason M. O'Kane and Ioannis Rekleitis}
}

@inproceedings{bearings,
   title={Underwater Surveying via Bearing Only Cooperative Localization},
   url={http://dx.doi.org/10.1109/IROS.2018.8593431},
   DOI={10.1109/iros.2018.8593431},
   booktitle={2018 IEEE/RSJ International Conference on Intelligent Robots and Systems (IROS)},
   publisher={IEEE},
   author={Damron, Hunter and Li, Alberto Quattrini and Rekleitis, Ioannis},
   year={2018},
   month=oct, pages={3957–3963} }
\bibliographystyle{plain}

\end{document}